# Computing the Spatial Probability of Inclusion inside Partial Contours for Computer Vision Applications

Dominique Beaini[a], Sofiane Achiche[a], Fabrice Nonez[b], Maxime Raison[a]

[a]Polytechnique Montreal, 2900 Edouard Montpetit Blvd, Montreal, H3T 1J4, Canada
[b]University of Montreal, 2900 Edouard Montpetit Blvd, Montreal, H3T 1J4, Canada

## Abstract

In Computer Vision, edge detection is one of the favored approaches for feature and object detection in images since it provides information about their objects' boundaries. Other region-based approaches use probabilistic analysis such as clustering and Markov random fields, but those methods cannot be used to analyze edges and their interaction. In fact, only image segmentation can produce regions based on edges, but it requires thresholding by simply separating the regions into binary in-out information. Hence, there is currently a gap between edge-based and region-based algorithms, since edges cannot be used to study the properties of a region and vice versa. The objective of this paper is to present a novel spatial probability analysis that allows determining the probability of inclusion inside a set of partial contours (strokes). To answer this objective, we developed a new approach that uses electromagnetic convolutions and repulsion optimization to compute the required probabilities. Hence, it becomes possible to generate a continuous space of probability based only on the edge information, thus bridging the gap between the edge-based methods and the region-based methods. The developed method is consistent with the fundamental properties of inclusion probabilities and its results are validated by comparing an image with the probability-based estimation given by our algorithm. The method can also be generalized to take into consideration the intensity of the edges or to be used for 3D shapes. This is the first documented method that allows computing a space of probability based on interacting edges, which opens the path to broader applications such as image segmentation and contour completion.

## Nomenclature

| | |
|---|---|
| $t$ | Time for parametric functions |
| $t_{i,f}$ | Initial and final time |
| $S$ | Stroke, defined as a partial contour |
| $S_C$ | Circular stroke |
| $\beta^\pm$ | The starting angle of a symmetric $S$ |
| $\beta^\pm$ | The starting angle from one side of $S$, or the other side |
| $S_C^\pm$ | The sections of $S_C$ defined with $\beta^\pm$ |
| $s_i$ | A part of the stroke $S$ |
| $G_k$ | A possible group of different $s_i$ or $S$ |
| $P_S$ | Probability of being enclosed in $S$ |
| $P_S^\pm$ | The value of $P_S$ associated to $V_m^\pm$ |
| $N$ | Number of strokes |
| $W_S$ | Weighted probability |
| $w_S$ | Weighted probability of a sub-image |
| $\gamma$ | A specific point in space |
| $\gamma_{i,f}$ | The starting and ending points of $S$ |
| $R$ | The region inside a closed $S$ |
| $V_m$ | Magnetic potential |
| $V_m^{\pm}$ | Region where the $V_m$ is expected to be positive/negative |
| $\boldsymbol{E}$ | Electric field, being the gradient of $V_m$ |
| $I$ | Image with value 1 on each stroke, and 0 elsewhere |
| $\theta$ | Orientation matrix perpendicular to the strokes in $I$ |
| $P_e$ | Electric potential by a single monopole |
| $P_{dip}^{\theta}$ | Complex dipole potential |
| $F$ | Density correction factor of $I$ |
| $\Omega(f)$ | Variance of $f$ |
| $\subset$ | Subset operator, used to indicate $\gamma$ is inside $R$ |
| $\cap$ | Intersection (AND operator) |
| $\cup$ | Union (OR operator) |
| $\bigcup_i a_i$ | Union of all the elements $a_i$ |
| $\circ$ | Hadamard product (Element-wise multiplication) |
| $*$ | Convolution operator |
| $\pm^?$ | Either sum or subtraction, to be decided |

**Definitions and acronyms**

| | |
|---|---|
| **Path** | A function of time $S(t)$ that starts at position $S(t_i) = \gamma_i$ and ends at position $S(t_f) = \gamma_f$ |
| **Contour** | A closed path with only 1 intersection at points $S(t_i) = S(t_f)$ |
| **Stroke** | Part of a contour (previously referred as partial edge or partial contour), for time $t_i \leq t \leq t_f$ |
| **Edge** | Weight associated to the probability that a given pixel is at the boundary of 2 regions |
| **CAMERA-I** | Convolution Approach of Magnetic and Electric Repulsion to Analyse an Image |
| **PIIPE** | Probability of Inclusion Inside Partial Edges |

## 1. Introduction

Image analysis and understanding is a challenging subject in computer vision, since there is an infinity of different images and videos that can be processed. Hence, properly extracting information from an image is a difficult task that often requires heavy computation and complex methodologies [1, 2]. One possible approach for image analysis is using probabilistic algorithms that allow comparing different parts of an image with their respective characteristics, which can be used for texture understanding [3, 4], image segmentation and clustering [5–7] and machine learning [8]. They are also used by several researchers for probabilistic image construction based on Markov fields or deep learning [9–11], allowing to fill parts of the images that are missing and generate artificial images.

One distinction between the cited algorithms is that edge-based methods generate information in a 1D space composed of thin edges [1, 2, 12, 13], while the region-based methods generate information in a 2D space composed of pixels [1, 2, 6, 7, 14]. Currently, multiple existing methods group edges to generate closed regions [15–18], but they do not provide any spatial information about the pixels not belonging to a contour. This implies that they cannot be used jointly with other region-based methods. Hence, there is a need to develop a novel probabilistic algorithm that generates spatial information based on the edges of an image, since it will close a gap in image analysis and could therefore unlock new possibilities. The approach proposed in this paper differs from any other existing algorithm since it provides spatial information based only on thin edges, a unique feature that does not exist elsewhere in the literature. This feature can then be used in different computer vision algorithms, such as contour completion [15–18] and edge-based image segmentation [19, 20] or saliency [21, 22].

In this paper, 4 similar concepts are used, being a *path*, a *stroke*, a *contour* and an *edge*. It is therefore important to fully understand the distinction between them. The full definitions are given in "Definitions and acronyms", with the time $t$ used to define the progression of the parametric functions, where $t_i$ is the initial time and $t_f$ is the final time. In summary, a path is any function $S(t)$, a contour is any non-self-intersecting closed path, a stroke is any partial contour, and an edge is a weight associated to a pixel present at the boundary of 2 regions. An example of those concepts is presented in Figure 1.

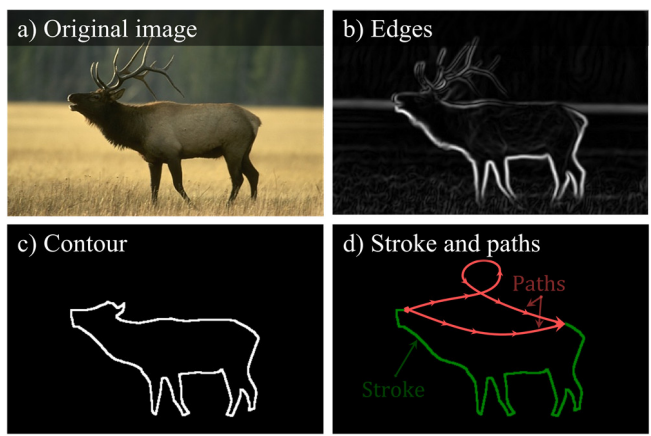

Figure 1. Definitions of different concepts. (a) Image of an elk from BSD500 dataset. (b) Edges computed using Sobel algorithm. (c) Contour of the elk. (d) Partial contour (stroke) along with 2 possible paths that close the stroke.

In our previous research work [23], we reported that electromagnetic (EM) convolutions allow to analyze different properties of a shape or a stroke. We demonstrated how the EM dipoles can be chosen to be invariant in regards to the size, the resolution and the orientation of a stroke, thus allowing its analysis. Also, it was confirmed that the EM kernels are robust to distortions and deformation [23, 24], which makes them ideal for the analysis of the general behavior of a complex stroke. Furthermore, we showed that the EM approach allows to generate information in the whole 2D space, based only on the 1D stroke. This allowed us to take into consideration the interaction between different strokes, their general concavity and to analyze the space between different strokes [23]. Improving the algorithms for stroke analysis can be useful in multiple applications, such as shape analysis [25, 26], object discovery [27–29] and object grasping [30, 31].

Building upon our previous research [23], the objective of the research work presented in this paper is to develop a new and improved method for computing the space probability of inclusion inside a partial contour using dipole electromagnetic convolutions [23], with the assumptions that any partial contour is meant to be closed and that different partial contours interact together. This paper will emphasis on developing the algorithm, but it will not present any application apart from the images used for exemplifying the mathematical concepts. Hence, it is the precursor of future application-focused work. The main objective is reached by completing the following steps:

1. Determine an analytical representation for computing the probability of being included inside partial contours, using a finite set of possible curves.
2. Generalize the results for a continuous space of probability using an uncountable set of circular curves.
3. Study the characteristics of the probabilities to ensure their consistency.
4. Demonstrate the equivalence between the space probabilities of step "2" and the computation of numerical magnetic convolutions.
5. Develop the algorithm to compute the space probability on complex images, where multiple shapes and contours are present.

The validation of the developed method will be carried out by showing how the partial contours can be used to generate an estimation of the original image which was used for edge detection [12, 32]. The approach is based on the premise that each edge should form a closed contour and uses this premise to compute the probability that each point in space is contained within the given contours. Hence, based only on their shape and their position, it can determine the regions of interaction and the partial contours that do not belong together. Thus, it differs fundamentally from any other probabilistic method in computer vision since it does not need information about color, texture, intensity, motion, etc.

The proposed technique is called PIIPE for Probability of Inclusion Inside Partial Edges, and it belongs to the general approach CAMERA-I [23, 24] (Convolution Approach of Magnetic and Electric Repulsion to Analyze an Image) developed in our laboratory at École Polytechnique de Montréal. Hence, the full name of the approach is CAMERA-I-PIIPE.

## 2. Computing the inclusion probabilities with circular paths

This section aims at understanding how to compute the probabilities that any point is enclosed within an open stroke, knowing that a single path should close the stroke. First, this section will justify that circular paths have the ideal characteristics for enclosing paths. Secondly, it will show how an infinite number of circular paths can be used to compute the probability of enclosure. Finally, the properties of the computed probabilities and their validity are analyzed.

### 2.1. The importance of subsets regions

This subsection presents the concept of computing the probability of inclusion for a stroke, which requires to consider different possible paths that close the given stroke. Although the most trivial path between 2 points at the extremities of the stroke is a straight line, the developed technique requires to consider different possible paths for the computation of the space of probabilities. This is because a single path to close the stroke will lead to only 2 possible values being "0" (outside the contour) and "1" (inside the contour). Hence, a space of probabilities other than "0" and "1" requires more possible paths.

To generate simple and intuitive paths, the paths between 2 points should be non-self-intersecting, convex and smooth, as discussed in more details in the appendix "B.1. Characteristics of the paths between 2 points". Then, it is possible to define a path $S_n$ that passes by the extremities $\gamma_{i,f}$ of a given stroke $S$. Therefore, if $S$ and $S_n$ do not intersect, it is then possible to define a region $R_n$ which is bounded by $S$ and $S_n$. This is shown in Figure 2, where the region $R_n$ contains the point $\gamma_{in}$ but excludes the point $\gamma_{out}$. A more rigorous definition of $R_n$ will be given at section "2.3 Intersecting circular arcs" in equation (9).

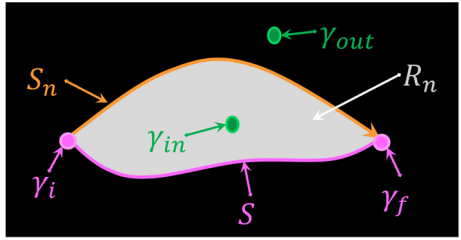

Figure 2. Example of a stroke $S$ between points $\gamma_{i,f}$, closed by a path $S_n$ to generate the region $R_n$ containing the point $\gamma_{in}$ but excluding $\gamma_{out}$.

The probability that a given point $\gamma_{in}$ is inside the region $R_n$ can be computed if we allow a finite number of regions $N_R$ that are partially bounded by $S$, where there is a smaller number $N_\gamma$ of regions $R_n$ that contain $\gamma_{in}$ ($\gamma_{in} \subset R_n$), versus the total number of regions $N_R$. Then, by assuming that each path $S_n$ is equiprobable, it is possible to compute the probability $P_S$ of being inside the stroke $S$ using equation (1).

$$P_S(\gamma \subset R_n) = \frac{N_\gamma}{N_R} \tag{1}$$

To compute the probabilities given by (1), it is required to find the values of $N_\gamma$ and $N_R$. To significantly reduce the complexity of the problem, we can choose the paths $S_n$ such that it does not intersect any other path $S_{m \neq n}$, except at the points $\gamma_i$ and $\gamma_f$ (noted $\gamma_{i,f}$). We also define $S_n^+$ and $S_n^-$, with each sign representing a path on a different side of $S$. The numbering variable $n^\pm$ and the angle $\beta_n^\pm$ are also defined according to the sign and numbering of $S_n^\pm$.

Therefore, if we suppose that a path $S_n^\pm$ does not intersect $S_{m \neq n}^\pm$, that it is associated to a starting angle $\beta_n^\pm$ (refer to Figure 3), and that each angle $\beta_n^\pm$ is smaller than the next angle $\beta_{n+1}^\pm$, than we can deduce that each region $R_n^\pm$ will be a subset of the region $R_{n-1}^\pm$. This relation is expressed in equation (2), with an arbitrary example presented in Figure 3 using $n^+ = [1, \ldots, 5]$ and $n^- = [1, 2]$.

$$\left. \begin{array}{c} S_n^\pm(t_n) \neq S_{m \neq n}^\pm(t_m) \, \forall \, \{t_n, t_m\} \\ \text{AND} \\ \beta_n^\pm < \beta_{n+1}^\pm \end{array} \right\} \Rightarrow R_n^\pm \subset R_{n-1}^\pm \tag{2}$$

Since the angles $\beta^+$ and $\beta^-$ have the same starting and ending points but in different directions, then the relationship between them is given by equation (3).

$$\beta^- = 2\pi - \beta^+ \tag{3}$$

For any non-infinite value of $N_R$, the value of $N_\gamma$ depends of the value of $n$ that respects the condition $\gamma \subset R_n$. For example in Figure 3 where $N_R = 7$, the probability $P_S$ of any point being within the region $R_n$ can be computed using equation (1), with the result shown in equation (4).

$$P_{S_{\text{example}}}(\gamma \subset R_n^\pm) = \frac{N_\gamma}{N_R} = \frac{n^\pm}{N_R} = \frac{n^\pm}{7} \tag{4}$$

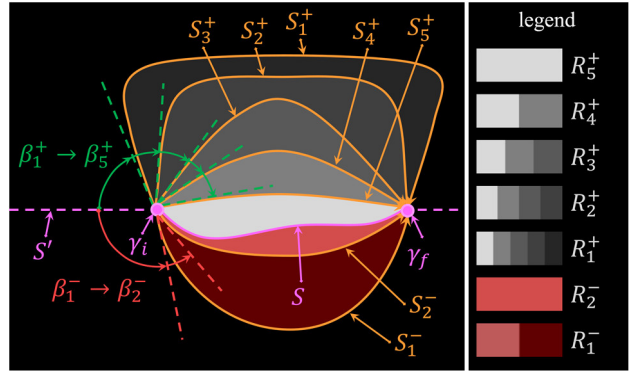

Figure 3. Example of 7 paths $S_n$ between points $\gamma_i$ and $\gamma_f$, with starting angles $\beta^+_{1\to5}$ and $\beta^-_{1\to2}$, such that $R_n$, the region between $S_n$ and $S$, is a subset of $R_{n+1}$

It is worth noting that since every region $R_n^\pm$ is a subset of $R_{n-1}^\pm$, it is possible compute the probabilities $P_S$ of belonging to the region $R_n$ in the case of a finite set of strokes using equation (1). Still, it is even more important in the case of an uncountable set of strokes, since it will allow generating a continuous space of probabilities. To generate such an uncountable set of strokes, one can define a stroke $S(\beta^\pm)$ for any angle $\beta^\pm = [0, 2\pi]$. Hence, there will be an infinite number of regions, meaning that the ratio in equation (1) will yield an indetermination. However, since there is a single curve $S^\pm$ associated to each angle, and since the regions $R_n^\pm$ are subsets of $R_{n-1}^\pm$, then the indetermination can be solved by replacing $N_R$ by the total span of $\beta^\pm$, and $N_\gamma$ by the span of $\beta^\pm$ such that $\gamma \subset R_n^\pm$. Therefore, the probabilities $P_S$ can be computed using equation (5), where $\beta_\gamma^\pm$ is the biggest angle that contains the point $\gamma$. Since $\beta_\gamma^\pm$ is bounded by 0 and $2\pi$, then the probability is also bounded by the inequality (6).

$$P_S(\gamma \subset R^\pm) = \frac{\text{range}\left(\beta^\pm(\gamma \subset R^\pm)\right)}{\text{range}(\beta^\pm)} = \frac{\beta_\gamma^\pm}{2\pi} \tag{5}$$

$$0 \leq P_S \leq 1 \tag{6}$$

## 2.2. Circular paths between 2 points

The previous section showed that it is possible to compute $P_S$ using equation (5) for an uncountable set of paths, without explaining how to generate such a set. Hence, this section will present how to generate a set using circular paths. Circular paths are ideal since they are smooth $C^\infty$, convex, symmetric and non-self-intersecting. Also, the set of circles passing by 2 constant points cover the entire 2D space, as discussed in more details in the appendix "B.2. Choosing the circle, rejecting the parabola".

An example of such circular path $S_C$ is given at Figure 4, where the only independent variables are $\beta$ and $x_0$, with $\beta$ being the starting angle and $x_0$ being the half-distance between the points $\gamma_{i,f}$. All the other variables, such as the radius, the area and the height of the circle, are dependent variables with the equations given in the appendix "B.3. Circular path parameters". The Cartesian equation of the circle is given at (7).

Let us note that the circle resulting of the angle $\beta^+$ is the same as the one resulting from the angle $\beta^- = \pi - \beta^+$, with $S_C^+(\beta^+)$ associated to one part of the circle, and $S_C^-(\beta^-)$ associated to the complementary part of the same circle (see Figure 4). Also, $S(\beta)$ is a set that contains an uncountable number of circles, since each angle $\beta$ represents a different circle. One could argue that it is easy to generalize the circular equation to an ellipse equation, but it violates the laws of electromagnetism discussed later in section "3 Computing the probabilities in an image using EM", as explained in more details in the appendix "C.1. Elliptical potentials and paths".

$$x_0^2 \csc^2 \beta = x^2 + (y - x_0 \cot \beta)^2 \tag{7}$$

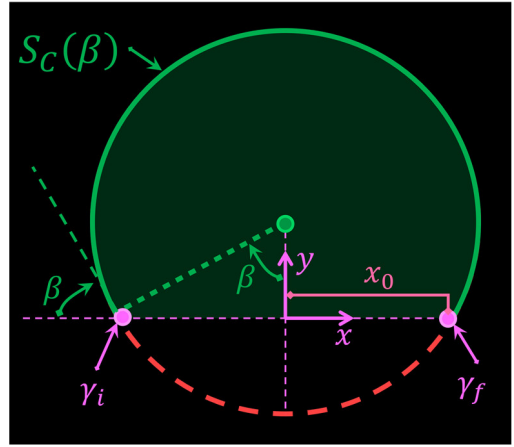

Figure 4. Example of a circular path between points $\gamma_i$ and $\gamma_f$, with a starting angle $\beta$

## 2.3. Intersecting circular arcs

The previous section presented the mathematical equations of a circle between 2 points, but it did not deal with the stroke $S$ that needs to be closed. This section will explain how to take it into consideration, and how to deal with multiple intersections between $S$ and $S_C$. This will allow to determine the region $R$ for any path $S_C$ and compute the probability $P_S$ for any point.

An example of a path $S$ closed by different circular paths $S_C(\beta_n^+)$ is shown at Figure 5, where the point $\gamma^+$ is at the boundary of $S_C(\beta_2^+)$ and well contained into $S_C(\beta_1^+)$. In that case, it is simple to compute the probability $P_S$ at any point along $S_C(\beta_{1,2}^+)$ using equation (5).

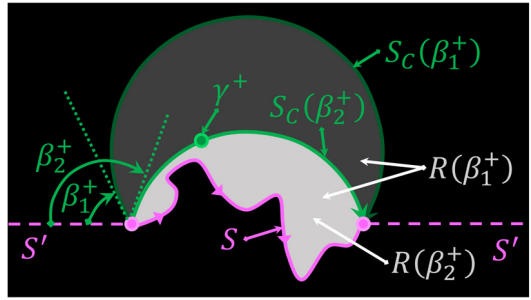

Figure 5. Example of 2 regions $R(\beta_{1,2}^+)$ formed by the closure of the path $S$ with the circular arcs $S_C(\beta_{1,2}^+)$

It becomes more complex to compute $P_S$ when there are intersections between $S$ and $S_C$ at the point $\gamma_\times$ in Figure 6, since it is harder to determine where is the region $R$. Such intersections will happen with any stroke $S$, except if $S$ is a circle with the same parameters as $S_C$. Therefore, it is important to be able to deal with such possibilities. In Figure 6, we can observe that the region $R$, which contains both points $\gamma^\pm$, can be defined as the region between $S$ and $S_C$, with $P_S(\gamma^+)$ associated to the angle $\beta^+$ and $P_S(\gamma^-)$ associated to the angle $\beta^-$. However, such definition does not hold well for non-trivial intersections.

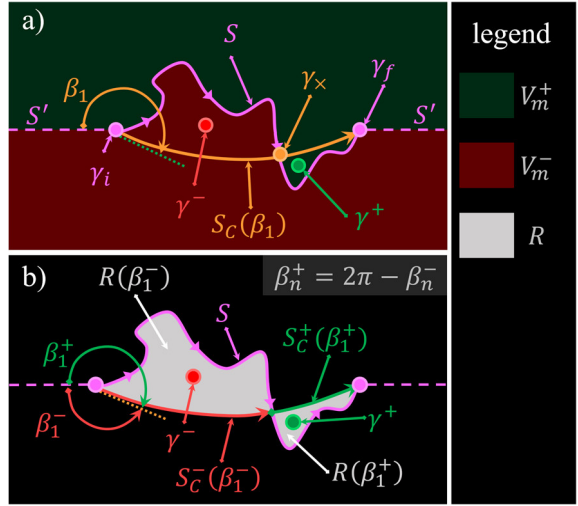

Figure 6. Example of (a) a stroke $S$ that intersect a circular arc $S_C$ at the point $\gamma_\times$. (b) The region inside the closure of the stroke $S$ with the sub-paths $S_C^-$ and $S_C^+$.

An example of complex intersection is given at Figure 7, where it is not intuitively clear which region should be counted inside or outside the grayed region $R$. To solve this problem, let's consider the infinite stroke $S_X'(t)$ as the continuation of the stroke $S_X$ along the line $\gamma_i \to \gamma_f$, as given by equation (8), where $S_X$ represents either $S$ or $S_C$. Also, to avoid unnecessary complications, we will assume that $S_X'$ is not self-intersecting. In that case, $S_X'$ separates the space in 2 half-spaces.

$$S_X'(t) = \begin{cases} S_X(t) & t_i \le t \le t_f \\ \dfrac{t - t_f}{t_i - t_f}\gamma_i + \dfrac{t - t_i}{t_f - t_i}\gamma_f & otherwise \end{cases} \tag{8}$$

For the half-spaces generated by $S_C$, we will define $R_C^\pm$ as the half-space containing $y \to \pm\infty$. For the half-spaces generated by $S$, we will define $V_m^\pm$ as the half space containing $y \to \pm\infty$. Then, the region $R$ will be defined by the region resulting of the logical equation (9), with an example depicted at Figure 7.

$$R = (V_m^+ \cap R_C^-) \cup (V_m^- \cap R_C^+) \tag{9}$$

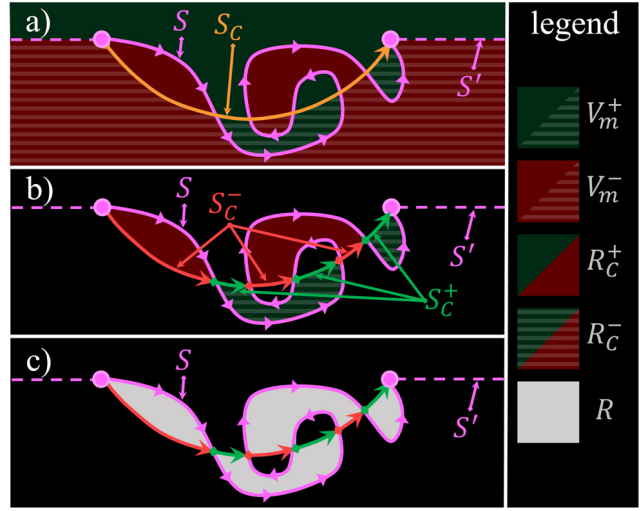

Figure 7. Example of a (a) A complex intersection between $S$ and $S_C$. (b) The regions that do not respect the logical operation are eliminated. (c) The region $R$ is formed with a union of the remaining regions.

Using the inside region definition of equation (9), it is possible to conclude that the probability $P_S$ of any point being inside $S$ is given by equation (5), where $\beta_\gamma^\pm$ is the angle that generates the elliptical arc $S_C^\pm$ that passes through the point $\gamma^\pm$. Hence, for any point in the region $V_m^+$, the value of $\beta^+$ is used in equation (5), while for any point in the region $V_m^-$, the value of $\beta^-$ is used.

## 2.4. Characteristics of the probabilities

The previous sections showed how to determine which points are inside the region $R$, and how to compute the probabilities using only the starting angle $\beta$. However, they must respect some basic properties in order to be mathematically valid, which will be the main focus of this subsection. It was already demonstrated that equation (5) respects the laws of probabilities with $P_S = [0, 1]$, since it respects the inequality (6). This section focuses on the analysis of other properties, such as certainty of inclusion/exclusion and complementarity, by exploring the mathematical boundaries of the model.

One boundary condition of the proposed mathematical model is that any point $\gamma_\infty$ infinitely far from $S$ must respect the equation (10). Since the only way for a circular path to reach a point infinitely far is when $\beta = 0$, then using equation (5) with $\beta = 0$ leads to equation (10).

$$P_S(\gamma_\infty) = 0 \tag{10}$$

Other characteristics can be studied at the boundary condition where $\gamma_S$ is defined as a point infinitely near $S$. We can choose a point $S_i$ on $S$, with a vector $\vec{v}$ perpendicular to $S$ at point $S_o$, as depicted in Figure 8 (a). Then, we can define the points $\gamma^\pm$ as 2 points situated at opposite side of $S$ at a perpendicular distance, as depicted in Figure 8 (a) and in equation (11). The point $\gamma_{S_n}^\pm$ is defined by the mathematical limit when the distance approaches 0 in equation (12). By computing the probabilities of $\gamma_S^+$ and using the equations (3) and (5), as seen in equation (13), we can find the property of complementarity presented at equation (14), with some visual examples at Figure 8 (b). This complementarity is required for the probabilities to make sense, since it means that the point $\gamma_S^+$ is inside $S$ only when $\gamma_S^-$ is outside $S$, and vice-versa.

$$\gamma^\pm = S_i \pm \vec{v} \cdot t \tag{11}$$

$$\gamma_{S_i}^{\pm} = \lim_{t \to 0} \gamma^{\pm} \tag{12}$$

$$P_S(\gamma_{S_i}^+) = \frac{\beta^+}{2\pi} = \frac{2\pi - \beta^-}{2\pi} = 1 - P_S(\gamma_{S_i}^-) \tag{13}$$

$$P_S(\gamma_{S_i}^+) + P_S(\gamma_{S_i}^-) = 1 \tag{14}$$

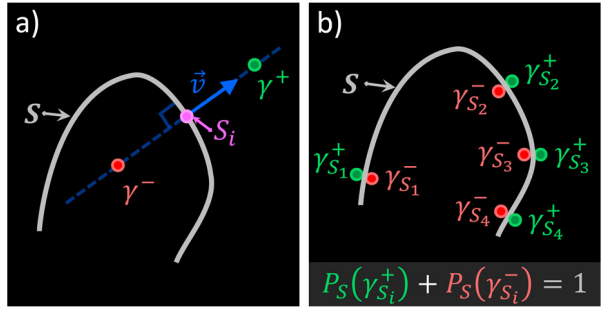

Figure 8. Complementarity of the enclosure probability across the path $S$. (a) 2 points at opposed side of $S$. (b) Multiple complementary points, at opposed sides of $S$, but with an infinitesimal distance.

Another important characteristic of the probabilities is that $P_S$ should have a value of 1 everywhere inside a closed contour, and a value 0 everywhere outside it. A closed contour can be viewed as any stroke where $\gamma_i$ and $\gamma_f$ are coincident, meaning that all the circles $S(\beta \neq \pi)$ have a null radius. Hence, the equation (10) forces any point outside the shape to have a value of $P_S = 0$, since $P_S(\gamma_\infty) = 0$ and since there are no circular paths $S(\beta)$ to change its value when $\gamma$ approaches the closed contour. Therefore, $P_S$ is constant both inside and outside the closed contour and varies only at its boundaries. Hence, using equation (14) with $P_S(\gamma^-) = 0$ allows to demonstrate that $P_S(\gamma^+) = 1$.

Finally, if we suppose that $S$ is the stroke formed by multiple sub-strokes $s_i$, then we need that the probability $P_S$ computed on the stroke $S$ to be the same as the combined probabilities $P_{s_i}$ computed on each sub-stroke. To make the problem easily solvable, we need to consider that the probability $P_S$ be the sum or subtraction of all $P_{s_i}$, as described by equation (15). The operator "$\pm^?$" means that the sign is chosen as positive or negative such that $P_S$ respects the previously stated conditions, and will be discussed in section "3.2.1 Repulsion optimization".

$$P_S = \sum_{i=1}^{n} \pm^? P_{s_i}, \quad \text{iff } S = \bigcup_{i=1}^{n} s_i \tag{15}$$

In summary, there are 5 fundamental properties presented at Table 1 that must be respected for the probabilities to be consistent with the mathematics and the boundary conditions.

Table 1. List of fundamental properties for the consistency of the probabilities

| # | Properties | Description |
|---|---|---|
| **1** | Laws of probability | Each probability is bounded by equation (6). |
| **2** | Certainty of exclusion | Any point $\gamma_\infty$ at an infinite distance of $\gamma_{i,f}$ has a value of $P_S = 0$ (equation (10)). |
| **3** | Complementarity | $P_S$ must be complementary on 2 points at each side of a stroke, when the distance between those points is infinitesimal (equation (14)). |
| **4** | Combination of probabilities | $P_S$ is the sum or subtraction of the probabilities given by each sub-stroke (equation (15)), such that conditions 1 and 2 are respected. |
| **5** | Certainty of inclusion | $P_S$ must be 1 inside a closed stroke, and 0 outside it. Proven with properties #1,2,3. |

## 3. Computing the probabilities in an image using EM

Although we explored the theoretical possibility of computing the probabilities of inclusion, this section is required to present how the EM potentials of dipoles allow to generate all those probabilities using mathematical convolutions in an image. First, it demonstrates that the equipotential lines are circular when the bi-dimensional dipoles are perpendicular to the stroke, and that they are related to the paths $S_C$. Then, it shows how multiple potentials can be combined to form a space of probability of belonging to any stroke $P_S$, for any pixel in an image composed of multiple non-trivial strokes.

### 3.1. Circular paths transform using EM potential

This subsection demonstrates that the dipole potential allows to generate the space of all possible circles, and to directly determine the value of $P_S$ on a single stroke $S$, using a magnetic convolution. Hence, the complexity of analyzing an infinite subset of circles and their intersections with $S$ will be greatly simplified, thanks to its mathematical equivalence with magnetic potentials.

### 3.1.1. EM convolutions

In order to compute EM potentials in an image, it is necessary to use convolutions to reduce computation time and ease the equations, as stated in previous work by Beaini et al. [23, 24]. The electric potential $P_e$ of a single charge in any universe of dimension $n$ is given by equation (16), where $\boldsymbol{r}$ is the Euclidean distance [23]. In a 2D image, the value of $n$ must be 2 to allow for conservation of energy and the use of Gauss theorem. Furthermore, it was shown that the potential of a dipole can be written as the complex potential given by the partial derivatives in equation (17) [23].

$$P_e = \begin{cases} |\boldsymbol{r}|^{\,2-n} & , \ n \geq 1, \ n \neq 2 \\ \ln|\boldsymbol{r}| & , \quad n = 2 \end{cases} \tag{16}$$

$$P_{dip}^{\theta} \approx \frac{\partial}{\partial x}(P_e) + i\,\frac{\partial}{\partial y}(P_e) \tag{17}$$

Furthermore, these EM potentials can be easily applied to an intensity image $I$ by using the convolution in equation (19) with the correction factor $F$ (18) [23, 24]. In the current paper, $I$ is the matrix with a value of 1 at the thin stroke and 0 elsewhere, and $\theta$ is direction of the stroke at any point in the matrix $I$.

$$F = \max(|\cos(\theta)|, |\sin(\theta)|)^{-1} \ \Rightarrow \ 1 \leq F \leq \sqrt{2} \tag{18}$$

$$V_m = \left( I \circ F \circ e^{i\theta} \right) * P_{dip}^{\theta} \tag{19}$$

### 3.1.2. Bi-dimensional EM potential on a line

The first step is to compute the EM potential that is generated by a line between 2 points, if the line is composed of a uniform density of dipoles perpendicular to its direction. This is illustrated at Figure 9, with the series of dipoles pointing in the $\hat{y}$ direction.

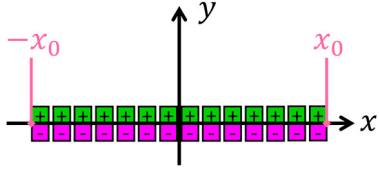

Figure 9. Line on the $x$ axis, composed of dipoles parallel to the $y$ axis.

To compute the potential generated by this line, we first must consider that the potential of a single dipole is the directional derivative of the monopole potential, with the directional derivative in the same direction as the dipole. Then, the contribution of all the dipoles can be taken using a definite integral with the boundaries being the positions $\pm x_0$, shifted by the $x$ position of each point [33, 34].

The total potential $V_m$ for the line depicted in Figure 9 is then given by equation (20), where $P_e$ is given by equation (16). By choosing $n = 2$, the potential $V_m$ is given by equation (21), with the result given at equation (22), where the values of $V_m$ are bounded by inequality (24) due to the previous arctangent.

$$V_m = \int_{x-x_0}^{x+x_0} \frac{\partial}{\partial y} P_e \ dx \tag{20}$$

$$V_m = \int_{x-x_0}^{x+x_0} \frac{\partial}{\partial y} \ln \left( \sqrt{x^2 + y^2} \right) dx \tag{21}$$

$$V_m = \mathrm{atan} \left( \frac{x + x_0}{y} \right) - \mathrm{atan} \left( \frac{x - x_0}{y} \right) \tag{22}$$

$$-2\pi \leq V_m \leq 2\pi \tag{23}$$

### 3.1.3. Circularity of the equipotential curves

The second step of the sub-section is to prove that the equipotential lines are circular. To prove it, we need to replace the inverse tangent in equation (22) by its complex form with the identity (24) and to define the variable $v$ with the expression (25), which yields to the equation (26). Then, by grouping the $x$ and $y$ together and by using trigonometric identities, we find the equation (27). The complete demonstration is presented in "Appendix D. Demonstration that equipotential lines are circular".

$$\mathrm{atan}(x) = \frac{i}{2} \ln \left( \frac{1 - ix}{1 + ix} \right) \tag{24}$$

$$v \equiv \exp(-2iV_m) \tag{25}$$

$$(v - 1)y^2 + (v - 1)x^2 + (v + 1)2x_0 yi - (v - 1)x_0^2 = 0 \qquad (26)$$

$$x_0^2 \csc^2 V_m = (y + x_0 \cot V_m)^2 + x^2, \qquad \{x, y, x_0\} \neq 0 \qquad (27)$$

An inspection of equation (27) shows that the equipotential lines are all circular, since each value of $V_m$ gives the equation of a circle. Furthermore, it is the same equation as the one for the circular path between 2 points given at (7), but with $V_m$ instead of $\beta$, which leads to equation (28), since the values are bounded by $\beta = [0, 2\pi]$ and $V_m = [-2\pi, 2\pi]$.

$$|V_m| = \beta \qquad (28)$$

### 3.1.4. Circular paths transform

The 3rd step is to be able to compute such potential on a stroke of any shape. The result of equation (28) means that, for a line $L$, the magnetic potential $V_m$ at any point $\gamma$ is equal to the starting angle $\beta$ of the circle that links the points $\gamma_{i,f}$ (both end of $L$) to the point $\gamma$. Hence, the computation of the probabilities $P_S$ at equation (5) becomes a simple computation of magnetic potential given by equation (29). Furthermore, it is possible to compute all the characteristics of equations (38), (39), (40) and (41) using $V_m$ instead of $\beta$ and $x_0$ as the half distance between $\gamma_i$ and $\gamma_f$.

$$P_S(\gamma \subset R) = \frac{|V_m|}{2\pi} \qquad (29)$$

The equation (29) is not useful if it can only be applied for a line. Hence, we need the equations (18) and (19) to compute the potential $V_m$ for any thin stroke in an image, since they allow the superposition of 2 perpendicular dipoles to create a dipole in any direction.

Using equations (17), (18) and (19), we can compute the circular equipotential lines for any stroke $S$. The reason why the equipotential lines stay circular is unknown, and a mathematical proof is beyond the scope of this paper. Nevertheless, it is observed numerically with many different shapes in Figure 10, where we can see that the expected circular equipotential (in white) match closely the magnetic equipotential lines (in green and pink). There are some numerical errors due mainly to a small error in the angle $\theta$, since the orientation of the strokes is estimated numerically. I call these equations "circular paths transform", since it allows to transform a 1D stroke into a 2D space of circular paths, with each circle passing through both ends of the stroke and its potential value corresponding to the starting angle $\beta$ of the circle. However, this allows an alternative way to compute the circular potential is given in the appendix "C.2. Convolution alternative".

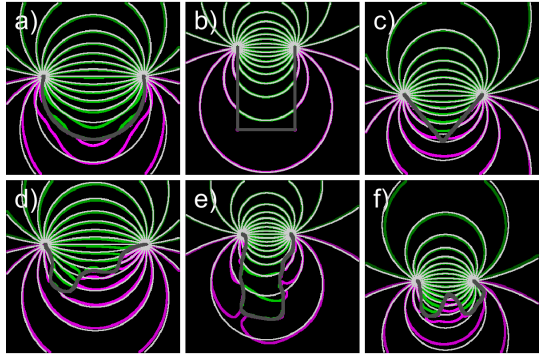

Figure 10. Example of equipotential lines of $V_m$ (green and pink) computed on 6 different strokes (dark grey), with the light grey lines being the perfectly circular equipotential lines of equations (27) and (28)

### 3.2. Scalar probability superposition

Computing the probability for a stroke can be useful, but it is usually required to compute the probabilities generated by multiple strokes in an image. Since the developed method relies on finding all the paths between the extremities of the stroke, then adding multiple strokes will require considering the paths between all those extremities. However, such problem becomes exponentially more complex with each new stroke that is added and yields to intersecting paths. This section explains how an understanding of magnetic potentials allows to simplify the computation and improve the results through repulsion optimization, double boundary detection and image splitting.

### 3.2.1. Repulsion optimization

There is one major problem when summing different potentials $V_m^i$, since the dipoles are aligned perpendicularly to the sub-strokes $s_i$. This means that the angle $\theta$ in equations (18) and (19) can be shifted by 180°, which will shift the sign of $V_m^i$ as seen in equation (30). Hence, there are 2 possible configurations for each sub-stroke in an image. This problem was raised previously with equation (15), where the sign "$\pm$?" was used to mention that it is either an addition or a subtraction, but without certainty.

$$\theta \rightarrow (\theta + \pi) \ \Rightarrow \ V_m^i \rightarrow -V_m^i \tag{30}$$

If there is a total of $n$ sub-strokes, then there should be a total of $2^n$ solutions, but the absolute value in equation (29) makes half the solutions redundant, meaning that there is a total of $2^{n-1}$ different solutions. However, there is only one solution that is consistent with equation (15), and it is the one where all the sides of $s_i$ are aligned according to their positive or negative sides. Hence, the magnetic repulsion must be maximized to be consistent with equation (15).

When the repulsion is maximized, there will be multiple regions that that form a constant potential as discussed in a previous paper by Beaini et al. [23]. At the boundary condition, a closed shape with all the dipoles aligned will generate 2 regions of constant potential with no gradient $\boldsymbol{E}$ except at the boundaries where $\boldsymbol{E}$ is high. In case the dipoles are not aligned, the value of $\boldsymbol{E}$ will vary smoothly between its minimum and maximum. Therefore, the distribution of $\boldsymbol{E}$ will be more split when the repulsion is maximized. Hence, we define the maximization parameter to be the variance $\Omega$ of $|\boldsymbol{E}|^2$ depicted in equation (31), meaning that $\Omega$ must be maximized to maximize the repulsion.

$$\Omega = Var(|\boldsymbol{E}|^2) \tag{31}$$

Since there are $2^{n-1}$ configurations, then it is preferable to use an optimization algorithm when $n$ is large to avoid long computing time. An algorithm that was developed and tested consist of creating a list $G$ which contains each individual index $i$, plus multiple groups of indices that are chosen according to their magnetic interaction. For example, the sub-strokes $s_i$ that connect with each other with a potential of $V_m > V_{th1}$ will form a group, those with a potential $V_m > V_{th2}$ will form another group.

Then, the potential $V_m^{G_k}$ of each element of $G$ are flipped and tested to see their impact on $\Omega$. If $\Omega$ is increased, then the elements of $G_k$ are permanently flipped. This algorithm is described in Figure 11 and was observed to work in most cases. If the number of elements are high, then the algorithm might end up in a local maximum. To avoid such problems, it can be used on different randomized initial orientations. Once each of them is optimized through the algorithm, the best solution must be chosen as the one with the lowest value of $\Omega$.

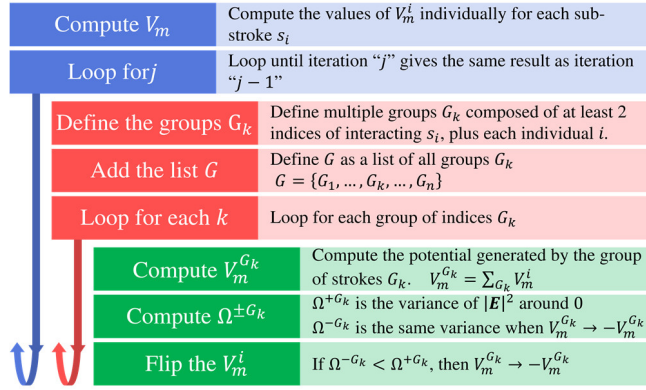

Figure 11. Algorithm used for repulsion optimization by flipping the magnetic orientation of each individual or group of sub-strokes $G_k$.

An example of such optimization is observed in the Figure 12, where the partial contours are extracted via the canny algorithm [32] with a high threshold. We can see that after the repulsion optimization, the high potentials $|V_m|$ are concentrated in the regions there are shapes, and the near zero potentials are between those shapes. It is to note that there are small regions where $|V_m| > 2\pi$, which are saturated in the Figure 12 and Figure 13.

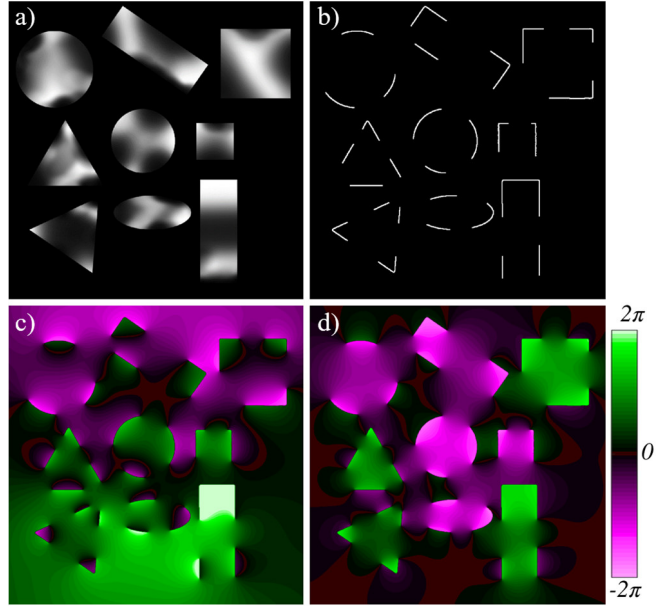

Figure 12. (a) Artificial image composed of different nearby shapes (b) Extracted partial contours using Canny [32]; (c) Resulting $V_m$ in the initial orientation; (d) resulting $V_m$ after the repulsion optimization.

The algorithm in Figure 11 was tested with the 28 strokes $s_i$ of Figure 12, and the result was compared to the brute force optimization that minimized $\Omega$ by testing the $2^{27}$ different configurations. The results were the same, but the computation time was around a $10^6$ times faster using the algorithm. This test was done with different images, including Figure 13, and the results were always the same, which shows that the algorithm converges to an optimal result.

### 3.2.2. Double boundaries

In some cases, a stroke will be at the boundaries of 2 different regions, which means that its contribution should be doubled to consider both regions. There are 2 equivalent ways of doing it, which are either to double the value of $F$ in equation (19) or to create a second stroke adjacent to the first one. An example with a few adjacent shapes is presented at Figure 13, where we can see the improvement of the potential when the double boundary is considered. One important improvement is the reduced potential between the shapes, so the high potential is mainly concentrated within the shapes. Another one is that the double boundary produces 2 clearer sides when it is considered, as seen by the circle and the triangle at the left. Furthermore, the 2 regions of the top rectangle are only distinguishable when the double boundaries are considered.

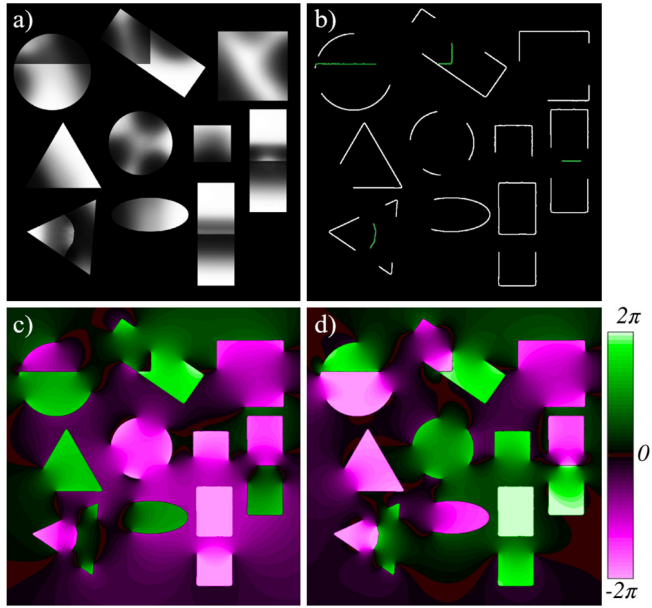

Figure 13. (a) Artificial image composed of different adjacent shapes (b) Extracted partial contours using Canny [32], with the double boundaries in green; (c) Resulting $V_m$ without the double boundary; (d) resulting $V_m$ with the double boundary.

### 3.2.3. Image splitting by attraction elimination

In the case where many different shapes are present in a single image, the repulsion optimization will still yield in some adjacent shapes that produce an attractive field between each other, since one will have a positive $V_m$, while the other will have a negative $V_m$. Since those shapes will be sure to not belong together, then they can be split into 2 new images that do not interact together. The algorithm to decide how to split them is presented at Figure 14, with the goal of reducing the initial potential image into multiple as much sub-images as possible, without loss of information. It is to note that this step is not mandatory since it increases the total computation time, although it usually improves the results. Also, some strokes might be in different sub-images, since they can belong to different groups. An example of the algorithm is presented in Figure 15.

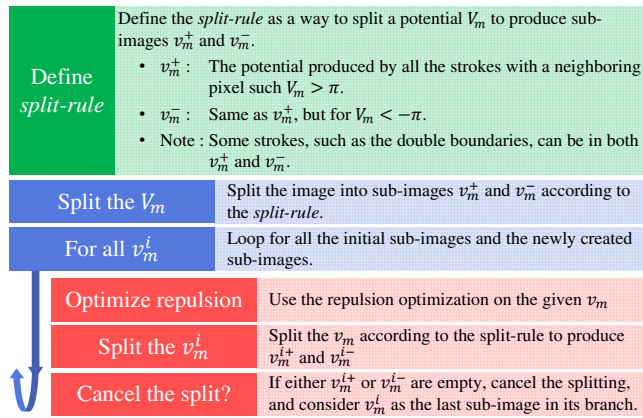

Figure 14. Algorithm used for the image splitting into multiple sub-images

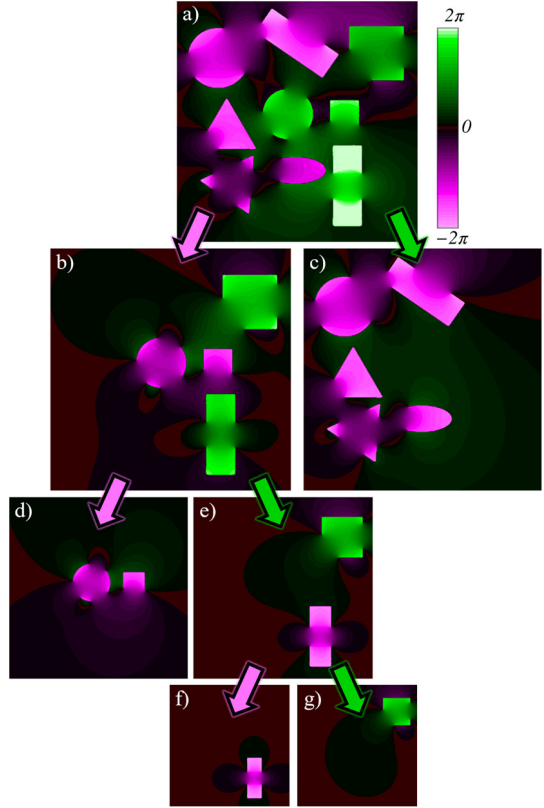

Figure 15. Example of image splitting process; (a, b, e) The temporary states of the splitting; (c, d, f, g) The final set of split potentials $v_m^i$

## 4. Important properties

With the knowledge of the previous sections, we know how to properly compute the probability $P_S$ using EM convolutions, but we did not discuss the interesting properties that arise. Hence, this section will cover some special features such as the weight adjustments, the equipotential line destination, the possibility of generalizing it in 3D and the information estimation.

### 4.1. About the probabilities

This subsection will focus on covering the closed shapes, the invalid probabilities and the possibility of adjusting the weight of each probability.

#### 4.1.1. Closed shapes

In some cases, a stroke may be already closed, which means that $P_S^+$ must be 1 inside $S$, and $P_S^-$ must be 0 outside $S$, as stated in the fundamental properties at Table 1. To prove it, we first use Gauss theorem, since it was demonstrated by Beaini et al. [23] that $V_m$ is constant both inside and outside of $S$ [33, 34]. Then, we know that the potential is null at a point $\gamma_\infty$ infinitely far, as seen in equation (22). Finally, we know from equation (43) that crossing the stroke leads to a potential variation $\Delta V_m = \pm 2\pi$, which means the value is 0 outside the stroke and $\pm 2\pi$ inside it. Hence, knowing from equation (29) that $P_S = |V_m|/2\pi$, we demonstrate that $P_S$ is 0 outside the shape and 1 inside it, which is consistent with the properties at Table 1.

Furthermore, we know from Beaini et al. [23] that Gauss theorem will only give a constant potential inside a shape if and only if the potential of a charge $P_e$ is proportional to the equation (16) [33, 34]. We also know that the probabilities are only consistent if we use dipoles that are perpendicular to the contour. Hence, we conclude that the potential $V_m = \left(I \circ F \circ e^{j\theta}\right) * P_{dip}^\theta$ given at equation (19) is the only possible potential that can be used for the computation of the probability of inclusion inside a stroke $S$, with $P_S = |V_m|/2\pi$ (29).

In summary, the developed method is believed to be the only possible way to compute $P_S$ using potential convolutions, since it is the only potential that will have a probability of 1 inside a closed stroke, and 0 outside it.

### 4.1.2. Invalid probabilities

In some other cases, the probabilities computed using equation (29) will be greater than 1, which is invalid mathematically. Most of those times, the probability will be in the interval $[1, 1.10]$, which is simply a numerical error. Most of those errors are one-off occurrences and can be solved by a median filter, while the rest can simply be rounded to the value 1. However, other cases will have a value that is in the interval $[1.10, 2]$, which happens when the given point is inside 2 shapes simultaneously. This is the result of an attraction instead of a repulsion or of a self-containing shape. Most of those problems are solved or reduced via the image splitting described in section "3.2.3 Image splitting by attraction elimination". However, the only way to permanently solve this problem is to saturate the values of $P_S$ for a maximum of 1, which is the approach used in this paper.

### 4.1.3. Weight adjustments

The proposed method allows to compute the probability $P_S$ using equation (29), but only if an equal weight is attributed to each of the circular equipotential. As it was discussed in section "4.1.1 Closed shapes", it is impossible to change the potential to add more weight for the shortest equipotential. However, it is possible to weight the probability $P_S$ by using a smooth-step function to obtain a weighted probability $W_S$ in equation (32), which is based on the Hermite polynomials and is valid for any value of $P_S = [0, 1]$ [35]. An example of the smooth-step function for $K = 2$ is given in equation (33).

A weight function will only work it is bounded by $[0, 1]$, strictly increasing and antisymmetric around $P_S = 0.5$. Since the smooth-step function respects those conditions, then it respects all the properties required for the probabilities to stay consistent with the fundamental properties given in Table 1.

In the case described in section "3.2.3 Image splitting by attraction elimination", it was explained that the probabilities will be better if the potential image is split into multiple sub-images. In that case, the total weighted probability $W_S$ is considered as the maximum value of all the weights of the sub-images $w_S^i$, as described in equation (34). Although equation (34) is not consistent with Table 1, it allows to determine what is the maximum probability of belonging inside a shape, which is still a relevant information. Otherwise, we can still access all the $w_S^i$ individually.

$$W_S = P_S^{K+1} \sum_{k=0}^{K} \binom{K+k}{k} \binom{2K+1}{K-k} (-P_S)^k \tag{32}$$

$$W_{S(K=2)} = 6P_S^5 - 15P_S^4 + 10P_S^3 \tag{33}$$

$$W_S = max(w_S^i) \tag{34}$$

## 4.2. Additional features

This subsection will cover other additional features that can be obtained by the $P_S$ or $V_m$, but without discussing them thoroughly. Those features include the equipotential line destinations, the possibility of computing uncertain partial contours and the possibility of analyzing 3D shapes.

### 4.2.1. Equipotential lines destinations

An interesting fact to note about the equipotential lines is that they always seem to pass through the extremities of the strokes, even when the image is complex, such as Figure 12 and Figure 13. In those images, we can see that only a few equipotential lines avoid the extremities, and it happens near the corners where the numerical error is higher. Also, some equipotential lines will cross the strokes and will be subject to the transformation at equation (43), but they will eventually reach the extremities. This fact means that, by using a variable threshold value on the potential, it is possible to obtain different hypothetical shapes that are formed by the given strokes.

### 4.2.2. Uncertain partial contour

In some cases, a part of a partial contour might not be certain to be an actual contour, and setting its stroke value to either 0 or 1 according to equation (19) might not be the best option. In that case, the matrix $I$ which is usually composed of 0 and 1, can be changed to be any real value bounded by 0 and 1. This will be equivalent of reducing the weight associated to the specific stroke. For example, the $P_S$ of a closed stroke with a value of 0.7 will be 0.7 inside it, and 0 outside it.

### 4.2.3. Probability analysis for 3D shapes

The work from the current paper can also be generalized for 3D partial surface $S_3$, where the proposed method would be able to compute the probability of belonging inside the solid. To do so, we need to use the equation (17) with a value of $n = 3$ and replace the factor $2\pi$ in equation (29) by the factor $4\pi$. Furthermore, the equations (18) and (19) need to be changed to consider 2 angles $\theta$ and $\phi$ to take into account the 3D orientation, such that each voxel in $I$ will have an orientation perpendicular to the surface at this point.

Using the equations in 3D will not produce circular shapes anymore, but complex 3D shapes. However, this does not impact the ability to compute the probabilities, since the results will still be consistent with the properties of Table 1, if the word "stroke" is replaced by "surface". Furthermore, in the boundary condition where the surface $S_3(x, y, z)$ is independent of $z$, then the computed probability $P_{S_3}$ in any $xy$-plane will produce circular equipotential lines, and $P_{S_3}$ will be the same as the probability $P_S$ computed with $n = 2$ on a stroke $S$, where $S = S_3$.

## 4.3. 3D information estimation from 1D strokes

Another aspect of the proposed approach is that it allows estimating the original image based only on the information available with the partial strokes, which is impressive since the strokes are 1D information, while an image is 3D information.

In fact, the image $I$ composed of the strokes $S$ represent 1D information, since the strokes are thin, and their value is either 0 or 1. However, the computation of $W_S$ using equation (33) generates 3D information, since it fills all the pixels in the image with a value in the range $[0, 1]$. Hence, a surprising characteristic of CAMERA-I-PIIPE is that the probabilistic reconstruction allows estimating the original 3D image (height, width and intensity) using only the 1D partial contours, as seen in Figure 16.

Although it is impossible to obtain the same image since most information is lost by taking the partial contours, the estimated results are extremely similar both in shape and in intensity to the original image.

Hence, Figure 16 shows that the probability computation is consistent with the expectations that $W_S$ allows estimating the original image using only its partial contours.

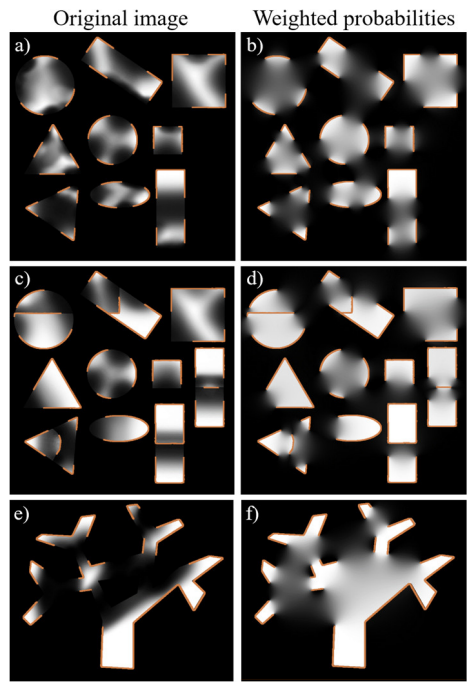

Figure 16. Comparison of the original images with the probabilistic reconstruction. (a, c, e) Original synthetic image composed of different shapes with the partial contours (orange); (b, d, f) Probabilistic weighted reconstruction based on the partial contours (orange).

## 5. Conclusion

The work presented in this paper detailed the development the CAMERA-I-PIIPE method, which allows to compute a spatial probability of inclusion $P_S$ according to initial partial contours. To do so, it explained how we can use an uncountable set of subset paths to $P_S$, called and how to generate such a set using all the possible circular paths via a simple potential convolution. Then, it showed how the magnetized contours can be manipulated to compute $P_S$ on complex images with multiple contours. Finally, different features were studied, such as the double boundaries, the weight adjustment technique, the uncertain edges and the information estimation.

This paper is a precursor to numerous possible studies for computer vision applications, since it created a novel approach that generates a space of probabilities based only on partial contours. For the first time, it is possible to directly combine contour information and region information for image processing. A continuation of this work could focus on developing specific applications in different computer vision fields such as saliency, image segmentation and contour completion. For now, most methods for these applications consider either the region information or the edge information. Hence, we expect that they will benefit from the promising results of the current work since it should allow to combine edge-based and region-based approaches together.

**Acknowledgment**

We would like to thank NSERC, through the discovery grant program RGPIN-2014-06289, and FRQNT/INTER for their financial support as well as MEDITIS (Biomedical technologies training program) through NSERC (FONCER) initiative.

## Appendix A. Supplementary nomenclature

This appendix presents the nomenclature that is used exclusively in the following appendices.

| | |
|---|---|
| $L$ | Length of $S$ |
| $A$ | Area of the circle $S_C$ |
| $\rho$ | Radius of the circle $S_C$ |
| $Y_{\max}$ | Height of the circle $S_C$ |
| $TAC$ | Total absolute curvature of $S$ |
| $\kappa$ | Local curvature of $S$ |
| $x, y$ | Horizontal and vertical position |

## Appendix B. Paths characteristics

### B.1. Characteristics of the paths between 2 points

This appendix will focus on the desired characteristics of a path that links two points together. Although the trivial path between those points is a simple straight line, the developed technique requires an infinite number of paths to compute the space of probabilities, not only the most optimal one.

For a path between 2 points noted $\gamma_i$ and $\gamma_f$, it is preferable to have a symmetrical path, since it is invariant to the swapping of $\gamma_i$ and $\gamma_f$. Examples of 4 different symmetric paths $S_{1-4}$ are shown at Figure 17, with a starting angle of $\beta$ at points $\gamma_{i,f}$ and a distance of $2 \cdot x_0$.

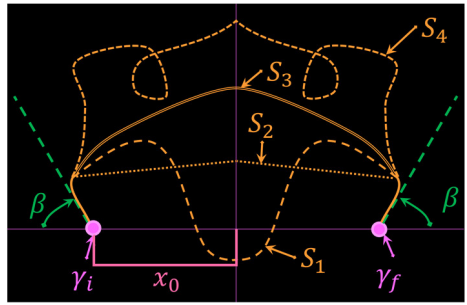

Figure 17. Example of different symmetric paths between points $\gamma_i$ and $\gamma_f$, with a starting angle $\beta$

For optimal paths, it is better to have a shorter length $L$ and a smaller total absolute curvature (TAC). The length measures the total distance, as given in equation (35) and the TAC is the integral of the curvature $\kappa$ in equation (36), with d$s$ is given at equation (37). For any closed curve, the following inequality is respected TAC $\geq 2\pi$, where it is only equal to $2\pi$ for the case of a convex curve.

$$L = \int_S ds \tag{35}$$

$$TAC = \int_S \kappa(s)\,ds\;, \qquad \kappa(s) = \frac{|\dot{x}\ddot{y} - \dot{y}\ddot{x}|}{(\dot{x}^2 + \dot{y}^2)^{3/2}} \tag{36}$$

$$ds = \sqrt{\dot{x}^2 + \dot{y}^2}\,dt \tag{37}$$

Another important characteristic of a path is its smoothness, noted $C^k$, where $k$ is the number of derivatives of the path that are continuous. The higher is the value of $k$, the smoother is the path.

With a quick inspection of Figure 17, it is easy to determine that an optimal path should not be self-intersecting, since it will pass by the same point more than once. Hence, the loop present in $S_4$ could simply be removed for a shorter path with a lower total absolute curvature. Also, the curve $S_1$ is concave, meaning that the TAC is not minimized. Finally, the curve $S_2$ is not smooth since its first derivative is not continuous. Therefore, the only curve in Figure 17 that respects all the criteria is $S_3$, as seen at Table 2.

Table 2 : Qualitative Comparison Between the Strokes Presented at Figure 17

| Stroke $S_n$ | Non-self-intersecting | Convex | Smooth |
|:---:|:---:|:---:|:---:|
| $S_1$ | ✓ | | |
| $S_2$ | ✓ | ✓ | |
| $S_3$ | ✓ | ✓ | ✓ |
| $S_4$ | | | |

## B.2. Choosing the circle, rejecting the parabola

The current appendix will explain why circular paths form optimal sets for this problem, which requires to create paths that passes through 2 points, with 2 defined starting angles $\beta$. This gives a total of 4 conditions on any non-symmetric path, but only 3 conditions on a symmetrical path (since the angle $\beta$ is symmetric).

If the path is chosen as a polynomial, then there would be an infinite number of possibilities for any polynomial of degree higher than 2 at any angle $\beta$. However, the equation (5) requires that there should be a single possible path per angle $\beta$, meaning that the only possible polynomial path is a parabola. The problem with parabolas is that the angle $\beta^+$ should be greater than $\pi/2$ for a path to exist between $\gamma_i$ and $\gamma_f$, otherwise they would diverge. Also, there is no path in the whole space where $\beta^\pm < \pi/2$, meaning that $P_S$ will be zero, which is not desired.

To solve those problems with the parabolas, we are forced to consider the non-polynomial paths which can respect the given criteria. One of the possibilities is the circle, since there is only a single circle that passes through 2 points with a given angle $\beta$, it is symmetric, non-self-intersecting, convex, and smooth $C^\infty$. Also, given 3 points, it is always possible to draw a circle that passes through all of them. If the 3 points are aligned, then it is possible to draw a circle of infinite radius. Therefore, the whole space will be covered, and there will be a circle for every angle $\beta = [0, 2\pi]$.

In summary, the parabola does not fit the required conditions well, while the circle fits them perfectly.

### *B.3. Circular path parameters*

The cartesian equation of the circular path $S_C$ is given at (7), with an illustration of all its parameters at Figure 18. From this equation, we can easily find the radius $\rho$ given at (38). Also, since the focus is only on the arc $S_C(\beta)$ seen at Figure 4, we can define the height between the $x$ axis and the top of $S_C$ as $Y_{max}$ given at equation (39). Furthermore, the length $L$ of $S_C$ is given by equation (40), and the area $A$ between the $x$ axis and the path $S_C$ is given by equation (41). No proof of these equations is provided since they can be demonstrated with basic trigonometry, and they can be tested for the boundary conditions at $\beta = \{0, \pi\}$ and for the half circle at $\beta = \pi/2$.

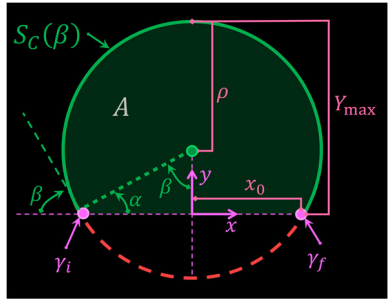

Figure 18. Example of a circular path between points $\gamma_i$ and $\gamma_f$, with a starting angle $\beta$

$$\rho = x_0 \csc \beta \tag{38}$$

$$Y_{max} = x_0 \cot \frac{\beta}{2} \tag{39}$$

$$L = \frac{2x_0(\pi - \beta)}{\sin \beta} \tag{40}$$

$$A = x_0^2 \left( \frac{\pi - \beta}{\sin^2 \beta} + \cot \beta \right) \tag{41}$$

## Appendix C. Electromagnetic potential

### *C.1. Elliptical potentials and paths*

It was previously discussed with equation (7) that each equipotential curve forms a perfect circle. It is easy to generalize it to any ellipse passing from the same points by using the transformation (42), where $b$ is the semi minor axis. However, this changes the values of the equations (38) to (41), which are outside the scope of the current paper.

Furthermore, such transformations do not obey Gauss law, the conservation of energy, or the diagonal superposition of dipoles of equation (17), meaning that the potential will not be constant inside a closed shape. Hence, they cannot be used for the computation of probabilities. Also, the equipotential lines will not be elliptical, unless the stroke where the potential is computed is a line, or unless we use the convolution alternative given in appendix "C.2. Convolution alternative".

$$y \to \frac{y}{b} \tag{42}$$

*C.2. Convolution alternative*

Another way to compute the circular potential without the convolution given in equation (19) is to use the equation (22) directly, with the coordinate system placed at the middle of the line between $\gamma_i$ and $\gamma_f$, and the $x$ axis pointing towards $\gamma_f$. Then, using the same definitions of $V_m^\pm$ given at equation (9), we transform the value of $V_m$ with equation (43). This allows to make sure that $V_m$ is positive in the region $V_m^+$, and negative otherwise.

Furthermore, using equations (22) along with the transformation (43) might be faster to compute than equation (19) since it does not require the use of convolutions, but it requires additional time to correctly identify the regions $V_m^\pm$ and additional time to process multiple strokes in the same image individually.

$$V_m \to \begin{cases} V_m - 2\pi, & V_m^+ \cap (V_m < 0) \\ 2\pi - V_m, & V_m^+ \cap (V_m > 0) \\ V_m, & \text{otherwise} \end{cases} \tag{43}$$

## Appendix D. Demonstration that equipotential lines are circular

Starting from $V_m$ given in equation (22) and using definition (25) and identity (24), the goal of this appendix is to demonstrate that the potential $V_m$ has circular equipotential lines. The demonstration is done by finding transforming equation (22) into the parametric equation of the circle given in (27). The full demonstration is given in Figure 19 below.

Initial equation

$$V_m = a\tan\left(\frac{x+x_o}{y}\right) - a\tan\left(\frac{x-x_o}{y}\right)$$

Identity 1: $\quad a\tan(X) = \frac{i}{2} \ln\left(\frac{1-ix}{1+ix}\right)$

Identity 2: $\quad \ln(X) - \ln(Y) = \ln\left(\frac{X}{Y}\right)$

$$\Rightarrow V_m = \frac{i}{2} \ln\left[\frac{\left(1 + \frac{i(x-x_o)}{y}\right)\left(1 - \frac{i(x+x_o)}{y}\right)}{\left(1 + \frac{i(x+x_o)}{y}\right)\left(1 - \frac{i(x-x_o)}{y}\right)}\right]$$

Defining $v \equiv e^{-2iV_m}$, $A \equiv \frac{x-x_o}{y}$, $B \equiv \frac{x+x_o}{y}$

$$\Rightarrow v = \frac{(1+Ai)(1-Bi)}{(1+Bi)(1-Ai)} = \frac{1 + (A-B)i + AB}{1 - (A-B)i + AB}$$

$$A - B = \frac{-2x_o}{y} \qquad AB = \frac{x^2 - x_o^2}{y^2}$$

$$\Rightarrow v = \frac{y^2\left(y^2 - 2x_o y i + x^2 - x_o^2\right)}{y^2\left(y^2 + 2x_o y i + x^2 - x_o^2\right)}$$

$$\Rightarrow v\left(y^2 + 2x_o y i + x^2 - x_o^2\right) = y^2 - 2x_o y i + x^2 - x_o^2$$

Intermediate form

$$\Rightarrow (v-1)y^2 + (v-1)x^2 + (v+1)2x_o y i - (v-1)x_o^2 = 0$$

$$(v-1)\left(y^2 + 2\frac{v+1}{v-1}ix_o y - \left(\frac{v+1}{v-1}\right)^2 x_o^2\right) = -(v-1)x^2 - \frac{(v+1)^2}{v-1}ix_o^2 + (v-1)x_o^2$$

$$\left(y + \frac{v+1}{v-1}x_o i\right)^2 = -x^2 + \left(i\frac{v+1}{v-1}\right)^2 x_o^2 + x_o^2$$

$$\left(y + \frac{v+1}{v-1}x_o i\right)^2 + x^2 = \left(i\frac{v+1}{v-1}\right)^2 x_o^2 + x_o^2$$

Identity 3: $\quad \cot(X) = -i\frac{e^{-iX} + e^{iX}}{e^{-iX} - e^{iX}}$

$$\Rightarrow \cot(V_m) = -i\frac{e^{-iV_m} + e^{iV_m}}{e^{-iV_m} - e^{iV_m}} = -i\frac{e^{-iV_m}\left(e^{-2iV_m} + 1\right)}{e^{-iV_m}\left(e^{-2iV_m} - 1\right)}$$

$$\Rightarrow \cot(V_m) = -i\frac{v+1}{v-1}$$

$$\left(y + x_o \cot(V_m)\right)^2 + x^2 = x_o^2\left(\cot^2(V_m) + 1\right)$$

$$\Rightarrow \left(y + x_o \cot(V_m)\right)^2 + x^2 = x_o^2 \csc^2(V_m) \quad \square$$

Figure 19. Demonstration that the potential $V_m$ has circular equipotentials